\pdfoutput=1

\documentclass[11pt]{article}

\usepackage[final]{acl}

\usepackage[toc,page,header]{appendix}
\usepackage{minitoc}

\usepackage{times}
\usepackage{latexsym}
\usepackage{multirow}
\usepackage{array}

\usepackage[T1]{fontenc}

\usepackage[utf8]{inputenc}
\usepackage{booktabs}

\usepackage{microtype}

\usepackage{inconsolata}

\usepackage{subcaption}
\usepackage{graphicx}
\usepackage{enumitem}
\usepackage{amsmath}

\usepackage{algorithm}
\usepackage{algpseudocode}
\usepackage{xcolor}
\usepackage{booktabs}
\usepackage{tabularx}

\title{SIMBA UQ: Similarity-Based Aggregation for Uncertainty Quantification in Large Language Models}

\author{
  \textbf{Debarun Bhattacharjya\textsuperscript{1}},
  \textbf{Balaji Ganesan\textsuperscript{1}},
  \textbf{Junkyu Lee\textsuperscript{1}},
  \textbf{Radu Marinescu\textsuperscript{1}}, \\
  \textbf{Katsiaryna Mirylenka\textsuperscript{2}},
  \textbf{Michael Glass\textsuperscript{1}},
  \textbf{Xiao Shou\textsuperscript{3}} \\
  \textsuperscript{1}IBM Research, 
  \textsuperscript{2}Zalando,
  \textsuperscript{3}Baylor University \\
    \{debarunb@us, bganesa1@in, Junkyu.Lee@, radu.marinescu@ie\}.ibm.com,\\
    katya.mirylenka@zalando.ch, mrglass@us.ibm.com, Xiao\_Shou@baylor.edu
}

\begin{document}
\maketitle
\begin{abstract}
When does a large language model (LLM) know what it does not know? Uncertainty quantification (UQ) provides measures of uncertainty, such as an estimate of the \emph{confidence} in an LLM's generated output, and is therefore increasingly recognized as a crucial component of trusted AI systems.  
\emph{Black-box} UQ methods do not require access to internal model information from the generating LLM and therefore have numerous real-world advantages, such as robustness to system changes, adaptability to choice of LLM,  reduced costs, and computational tractability. In this paper, we investigate the effectiveness of UQ techniques that are primarily but not necessarily entirely black-box, where the consistency between a generated output and other sampled generations is used as a proxy for confidence in its correctness. We propose a high-level non-verbalized \emph{similarity-based aggregation} framework that subsumes a broad swath of UQ approaches suitable for complex generative tasks, as well as introduce specific novel techniques from the framework that train confidence estimation models using small training sets. Through an empirical study with datasets spanning the diverse tasks of question answering, summarization, and text-to-SQL, we demonstrate that our proposed similarity-based methods can yield better calibrated confidences than baselines.
\end{abstract}

\section{Introduction}\label{sec:intro}

\emph{Uncertainty quantification} (UQ) approaches 
help inform 
reliability of model predictions and 
are therefore critical for deploying large language models (LLMs).
UQ refers to a broad suite of techniques that yield measures of uncertainty; here we are interested in assessing the \emph{confidence} of an LLM's generations for a user-specified task. 
While confidence is not always a well-defined quantity in the UQ literature, it typically captures some measure of faith in an LLM's response to a query as represented by a number between $0$ and $1$.
In this work, we consider tasks where there is a notion of whether an LLM's response to a user query is \emph{correct} or not, and interpret confidence of a response as the probability that it is correct.
This is in contrast to approaches that estimate the \emph{uncertainty} of LLM generations in response to a query, which measure the variability of the output.

An important category of UQ techniques are \emph{black-box} methods, which only
assume access to the model being used without requiring information such as the model parameters or even the token log probabilities.
Such techniques have numerous practical advantages as they are robust to the constantly evolving landscape of LLMs, 
computationally lightweight, and quickly deployable at inference time.
As a result, black-box UQ has become increasingly popular for tasks such as question answering~\citep{lin2024generating,cole2023selectively,manakul2023selfcheckgpt}. 
%
%

Much of the research on black-box UQ can be viewed as  \emph{consistency-based}, where the idea is to use the consistency between a generation and other sampled generations as a proxy for its confidence. 
The implicit underlying assumption behind such approaches is that when a generated response is more different from others, it is more likely to be incorrect, implying that responses that are consistently similar are more likely to be correct; this has been explored for various use cases involving self-consistency~\citep{mitchell2022enhancing,wangself,chen2024universal}. 
Recent work has referred to this assumption as the \emph{consistency hypothesis}, formalizing it through statistical tests and evaluation metrics, and empirically validating its prevalence across a suite of tasks and datasets~\citep{xiao2025consistency}.

\begin{figure*}
\centering
{\includegraphics[width=0.8\linewidth]{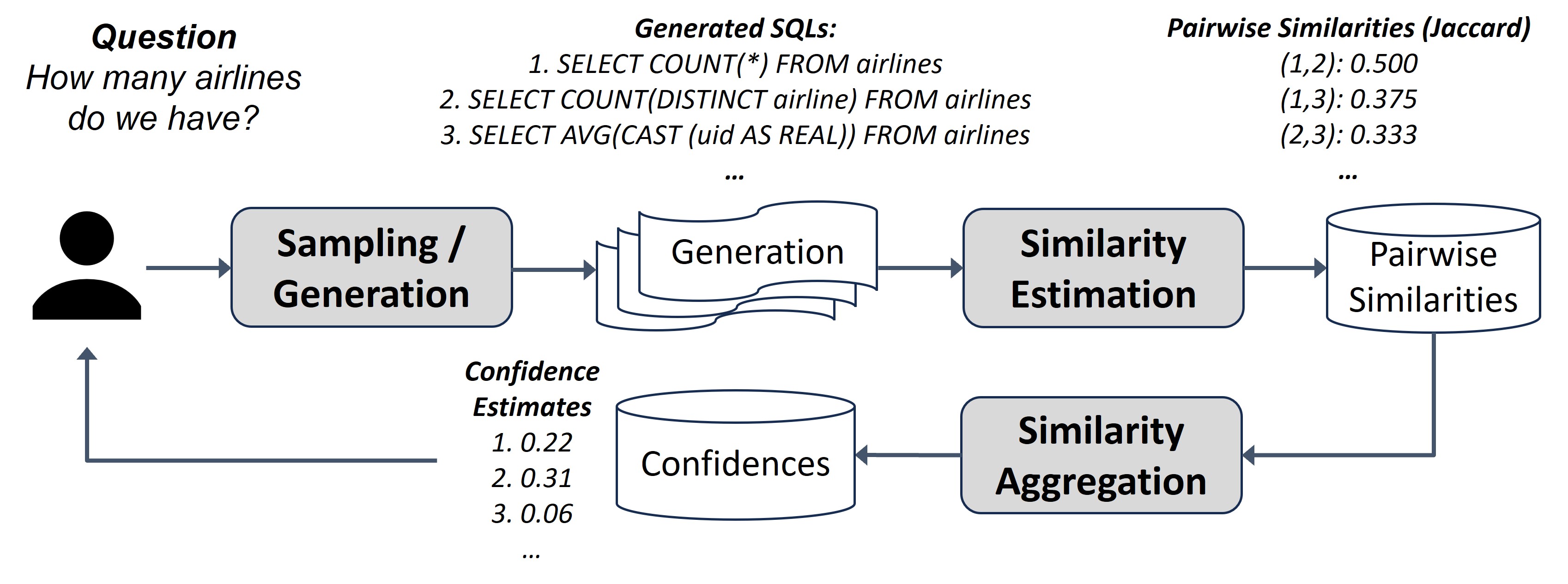} }
\caption{Framework for UQ yielding confidence estimates for generations from an LLM in response to a natural language query. An illustrative natural language query from the Spider dataset for the text-to-SQL task is shown, along with example outputs at each step of the pipeline. Jaccard is chosen as the similarity metric, and confidences are estimated using a proposed `aggregation by classification' method with random forests (described later).}
\label{fig:pipeline}
\end{figure*}

In this paper, we provide a fresh perspective on consistency-based approaches through the lens of aggregating pairwise similarities. 
Figure~\ref{fig:pipeline} outlines the procedure for a high-level framework that aims at exploiting the afore-mentioned consistency assumption for UQ. 
First, multiple outputs/samples are generated by the LLM through some sampling procedure. Pairwise similarities between samples can then be computed using any similarity metric of choice. Finally, these similarities are leveraged to provide confidence estimates for each generation of interest. Our methodological contributions are 
primarily in the third phase of Figure~\ref{fig:pipeline}. 
In particular, rather than viewing the third phase as clustering outputs like in closely related work~\citep{kuhn2023semantic,lin2024generating}, we propose \emph{similarity-based aggregation} as a framework for estimating confidences.
In contrast to aggregating verbalized confidences~\citep{xiong2024llms}, we aggregate pairwise similarities between generations, making our approach \emph{non-verbalized} and therefore avoiding some empirically observed concerns around potential overconfidence when asking LLMs for probabilities~\citep{hu2023prompting,xiong2024llms}.
Verbalized confidence aggregation is known to struggle for complex generative tasks. 

Confidence estimates from UQ can be evaluated in various ways depending on how they will be used by the system or end user. 
We are primarily interested in approaches yielding confidences that are well \emph{calibrated}, as gauged by how closely they align with the empirical accuracy of the predictions~\citep{murphy1967verification,dawid1982well}. We also evaluate our proposed approaches 
on metrics that gauge how confidence estimates benefit when used to select from a set of generations or predict whether a generation is correct or not.

Our \textbf{contributions} are summarized as follows:
\begin{itemize}[noitemsep,nolistsep,leftmargin=*]
\item We introduce a high-level similarity-based aggregation (SIMBA) framework, 
unifying and generalizing
various consistency-based UQ approaches by positioning them as different ways to represent (and possibly learn) 
confidence as a function of similarities with other generations.
\item We propose specific novel approaches from the framework, including 1) Bayesian aggregation that uses pairwise similarities as evidence, and 2) viewing confidence estimation as classification and obtaining the probability of a generation being correct 
using similarities 
as features.  
\item 
We conduct experiments using 9 datasets -- 3 each for question answering, summarization, and text-to-SQL tasks -- including ablations. 
Similarity-based aggregation methods are shown to perform well on all chosen evaluation metrics, 
particularly those measuring calibration error.
Importantly, results indicate that the methods perform well across short and long form generations, including when output is structured, like in SQL queries. 
\end{itemize}

\section{Related Work}

Procedures for UQ estimate measures like the variability or confidence of LLM outputs, and can be categorized as either white-box or black-box. White-box methods assume access to the LLM’s internal components, such as model weights, logits, or embeddings. In contrast, black-box methods rely only on outputs, inferring confidence through alternative means. 
An orthogonal distinction is between verbalized and non-verbalized methods, where the former type prompts an LLM to express uncertainty in natural language, such as using phrases (``I don’t know'' or ``most probably'') or  quantitative indicators (``low'' or ``50\%'' or ``90\%'').

\paragraph{White-Box Methods.}
Early work on calibration in deep learning and transformers laid the groundwork for using token-level probabilities -- derived from model output logits -- as a signal for estimating the reliability of model predictions~\citep{kuleshov2018accurate,ott2018analyzing,desai2020calibration}.
More recently, \citet{kuhn2023semantic} propose semantic entropy based clustering on multiple samples generated from the model and then estimating confidence estimates by summing the token-level probabilities in each cluster. 
The use of token-level probabilities for UQ has become a rapidly burgeoning area of research, driven by growing interest in the predictive confidence of LLMs~\citep{fadeeva,aichberger2024semantically,lin2024contextualized,vazhentsev2025token}.
\citet{kadavath2022language} suggest a verbalized method  where the LLM generates responses and then evaluates them as either True/False; the probability that the model assigns to the generated token determines confidence.

Other approaches consider the LLM's internal state such as embeddings and activation spaces. For instance, \citet{ren2023outofdistribution} fit the embeddings for both inputs and outputs in the training data to a Gaussian distribution, and estimate the model's confidence by computing the distance of the evaluated data pair from this Gaussian distribution. Some methods probe the model's attention layers to discriminate between correct and incorrect answers~\citep{kadavath2022language,burns2023,Li2023,azaria2023internal}. Although these methods provide insights into the model’s linguistic understanding, they require supervised training on specially annotated data.

\paragraph{Black-Box Methods.}
One strand of research considers verbalized black-box methods, such as using an LLM to evaluate the correctness of its own generated answers in a conversational agent scenario~\citep{mielke2022reducing}.
\citet{xiong2024llms} conduct an empirical study on UQ for reasoning tasks, showing that LLMs tend to be overconfident when verbalizing their own confidence in the correctness of the generated answers. 
Other related work includes fine-tuning GPT-3 to verbalize uncertainty associated with responses~\citep{lin2022teaching}. 

Many black-box methods use similarity between multiple generations given an input question, where common choices of metrics are natural language inference scores \citep{kuhn2023semantic} 
or embedding-based similarity~\citep{spuq,farquhar2024detecting}. 
Such similarity metrics can be used to 
extend clustering algorithms for uncertainty quantification of LLMs
\citep{kuhn2023semantic,ao2024css,da2024llm,jiang2024graph}.
In this line of work, one assumes that inconsistency in responses correlates with incorrect or hallucinatory generations. 
For instance, \citet{manakul2023selfcheckgpt} propose a simple sampling-based approach that uses consistency among generations to find potential hallucinations. \citet{lin2024generating} estimate uncertainty based on analysis of a similarity matrix between generations, such as through the sum of the eigen-values of the graph Laplacian, the degree matrix, and the eccentricity. 
\citet{farquhar2024detecting} and \citet{nikitin2024kernel} quantify uncertainty in LLM outputs using semantic similarity kernels 
to capture fine-grained variation among responses.
Recent methods have also explored combining white- and black-box UQ~\citep{chen-mueller-2024-quantifying,shrivastava2023}.

\paragraph{Comparison with Prior Work.}
Our proposed framework and corresponding approaches lie within the (consistency-based) non-verbalized UQ category and are primarily black-box (although they can be made white-box if needed, by choosing features arising from token probabilities).
While there is a growing body of work on UQ methods leveraging similarity, our proposed framework generalizes beyond  semantic similarity -- which has been the main focus in the closest related work -- by emphasizing the functional relation between similarities and confidence of generations. This is important because not all generative tasks need to be concerned about semantic similarity of output, e.g., code generation. Also, we note that our proposed supervised methods in the framework differ from concurrent efforts~\citep{liu2024uncertainty,ulmer-etal-2024-calibrating,yaldiz-etal-2025-design} in that: 1) they are also applicable when only similarities (and not token probabilities) are used as features and therefore can be entirely black-box, and 2) they are light-weight and thus easily deployable in real-world systems. 
We compare variations of the proposed aggregation methods for our extensive experimental study.

\section{Methodology}
\label{methodology}

We propose an overarching framework as well as specific techniques that provide confidence estimates, possibly using a limited amount of training data.
In this section, we first present some basic notation and assumptions, and then
describe the components of the workflow depicted in Figure~\ref{fig:pipeline}. 

\subsection{Basic Notation and Assumptions}

Consider an LLM generating output $y$ for some input query $x$.
We assume there is an associated ground truth output $y^*$ for input $x$ as well as a binary reward $r \in \{0,1\}$ from a reward function $r(x,y,y^*)$. Importantly, we assume there is a way to gauge whether any particular generation (with corresponding ground truth) is correct or not, i.e. whether the reward is $1$ or $0$; $Y^*(x)$ denotes the set of correct responses for $x$. This may not be possible in some situations, such as when humans disagree about class labels~\cite{baan-etal-2022-stop}.
For tasks such as open-ended question answering and summarization, we deem the reward to be $1$ if a text similarity metric (e.g. Rouge-L) between the ground truth and  generated output exceeds a predetermined threshold; this has also been assumed by related prior work~\citep{kuhn2023semantic,lin2024generating}.
For text-to-SQL, reward is $1$ if the generated and ground truth queries return the same result upon query execution on the underlying database.

\subsection{Sampling}

Consistency-based approaches begin with obtaining multiple generations $y_1, \cdots, y_m$ for an input $x$.  
In this work, we adopt a sampling approach where multiple samples are generated from multiple temperatures using the next-token probability distribution of the LLM.
While there are other means of generating diverse samples~\citep{spuq}, our focus is primarily on leveraging temperature.
Varying temperature has been empirically shown to provide ample opportunity for assessing whether response variability is present, which is needed by consistency-based methods to help distinguish correct from incorrect responses in complex generations~\citep{zhu2024hot,xiao2025consistency}.

In practice, one may be interested in confidence estimates for only a subset of the generations, such as the ones most likely to be correct. Generations at higher temperatures could therefore be used merely for better estimating those at lower temperatures.

\subsection{Computing Pairwise Similarities}

After generating samples, consistency-based approaches rely on access to a similarity metric with which one can compute pairwise similarities
$s(y_i, y_j)$ for all sample pairs, assumed to lie in the interval $[0,1]$.
As shorthand, we denote the similarities between the $i^{th}$ generation and other generations as $\mathbf{s}_i = s_{i,1}, .., s_{i, i-1}, s_{i, i+1}, .., s_{i, m}$, where $s_{i, k} = s(y_i, y_k)$ is the similarity between samples $y_i$ and $y_k$.
For our experiments, we consider metrics that treat samples as general text or sets of tokens, such as the Jaccard coefficient and variations of ROUGE such as Rouge-1 and Rouge-L.
We note that any similarity metric can be used, but in our experience, metrics that treat generations as sets of tokens are most suitable for consistency-based UQ. Future work could explore adapting learnable similarity functions over higher level concepts, such as those using graph neural networks for complex entity matching~\citep{krivosheev2021business}.

\subsection{Similarity Aggregation}

The final phase of the pipeline relies on leveraging pairwise similarities for UQ.
Recall that the underlying assumption behind consistency-based approaches is that correct generations are more similar to other generations than incorrect ones. 
An aggregated similarity between a generation and  others therefore acts as a proxy for correctness.

We present a simple yet broad perspective on consistency-based approaches that is applicable to any generative task. Rather than clustering generations such as around semantic equivalence~\citep{kuhn2023semantic}, 
the confidence for sample $y_i$ can be estimated using a suitable aggregation: 
$c_i = f(\mathbf{s}_i)$, where $\mathbf{s}_i$ denotes pairwise similarities between $y_i$ and other samples. 
A deterministic function $f(\cdot)$ implies that identical generations yield identical confidences for the same sample set, which is a desirable property.
Furthermore, $f(\cdot)$ should reflect the underlying hypothesis around consistency-based methods, which is that more consistency is expected for correct answers. We propose the following $3$ categories, each differing in how 
aggregation function $f(\cdot)$ is selected and/or learned. 

\textbf{Simple Aggregation.}
A simple approach is to find an aggregate distance between $y_i$ and other generations, $\bar{d}_i = g(d_{i,1}, 
\cdots, d_{i,m})$ where distance
$d_{i,k} = 1-s_{i,k}$, and compute $c_i = 1 - \bar{d}_i$ since the aggregate distance lies in $[0,1]$. The rationale is that the consistency assumption suggests that a generation further removed from others is more likely to be incorrect.
While any form of aggregation $g(\cdot)$ is possible, we use the 
arithmetic mean for experiments, 
which simplifies as
$c_i = \bar{\mathbf{s}}_i$.
This form of aggregation is mathematically equivalent to the spectral clustering by degree approach in \citet{lin2024generating} and therefore treated as a baseline for experiments. We show later that this performs reasonably well on some (but not all) UQ metrics.


\textbf{Bayesian Aggregation.}
We propose a Bayesian form of aggregation that updates beliefs about confidence using similarities as evidence. Specifically, we compute the posterior probability of $y_i$ being correct given similarities with other generations:
\begin{equation*}
P(y_i \in Y^*| \mathbf{s}_i) = \frac{p_0 \prod_{k \neq i} {\alpha_i}}{p_0 \prod_{k \neq i} {\alpha_i} + (1-p_0) \prod_{k \neq i} {\beta_i}},
\end{equation*}
where $p_0$ denotes the prior $P(y_i \in Y^*)$, 
$\mathbf{s}_i$ are pairwise similarities between $y_i$ and other generations,
$\alpha_i = P(s_{i,k} | y_i \in Y^*)$, $\beta_i = P(s_{i,k} | y_i \notin Y^*)$. The formula makes two important assumptions: 1) similarities in $\mathbf{s}_i$ depend only on whether $y_i$ is correct, and 2) similarities in $\mathbf{s}_i$ are conditionally independent. The first 
reflects the underlying assumption about the relation between consistency and correctness,
allowing for a less variable distribution if $y_i$ is correct, while the second is made for 
tractability.
Bayesian approaches are popular in related but different areas, such as modeling uncertainty in parameters of neural networks~\cite{gal16,maddox19}.

Note that this approach requires a small training set to learn the parameters of the probabilistic model.
For experiments, we assume Beta distributions for the conditional similarity distributions; this requires $5$ parameters to be learned -- prior  $p_0$ and $2$ parameters each for the $2$ Beta distributions.

\textbf{Aggregation by Classification.}
We also propose treating similarity aggregation as a classification task; specifically, we train a probabilistic classifier for whether a response is correct using supervised learning with features based on similarities. 
Denoting $\mathbf{s}^f_i$ as the set of similarity features for classifying correctness, 
confidence for generation $y_i$ is computed as:
$c_i = P(y_i \in Y^*) = f(\mathbf{s}^f_i)$.
This method can be generalized by also including other non-similarity features $\mathbf{o}^f_i$, such as the generative score from the LLM, in which case $c_i =  f(\mathbf{s}^f_i, 
\mathbf{o}^f_i)$.

For our experiments, we learn the function $f(\cdot)$ using a random forest as the probabilistic classifier, since it was observed to perform better than other methods like logistic regression. 
We compare variations of our proposed classification approach with different feature sets. 
(Details are provided later.) 

This is a simple yet powerful extension of simple aggregation where the function is learned using a small training set.
Both this approach and the Bayesian 
one are more likely to be effective when the sampling procedure for training is identical to that during test time. In practical applications, this is reasonably straightforward to control, assuming the training and test sets are not too dissimilar, and the data with similarity features that is needed for training a classifier can be easily compiled using a small labeled dataset with ground truth responses.


\section{Empirical Investigation}
\label{experiments}

\subsection{Experimental Setup}

We summarize our experimental setup in this subsection. Note that we restrict ourselves to using representative open-source LLMs for generation.

\paragraph{Datasets.}We consider the following datasets: 
\begin{itemize}[noitemsep,nolistsep,leftmargin=*]
\item \textbf{QA}: We consider the open-book dataset CoQA~\citep{reddy2019coqa}, the closed-book dataset TriviaQA~\citep{joshi2017triviaqa}, as well as the closed-book  dataset Natural Questions~\citep{kwiatkowski2019nq}.  
QA is widely studied in the literature on UQ for LLMs.
\item \textbf{Summarization}: For this task, we experiment with the following datasets: XSum~\citep{xsum}, SamSum~\citep{samsum}, and CNN Dailymail~\citep{cnndailymail}. 
Note that summarization typically results in longer form generations as compared to QA.
\item
\textbf{Text-to-SQL}: We consider the popular Spider benchmark~\citep{yu2018spider},  
Spider-Realistic~\citep{deng2021structure}, which is a more challenging version of Spider, 
and BIRD~\citep{li2024can}, a recent cross-domain benchmark covering many professional domains. Text-to-SQL with LLMs is an increasingly popular area of research for LLMs, with ongoing efforts to improve generation robustness and reliability through techniques like schema linking and advanced grounding~\citep{dragusin2025grounding}.
\end{itemize}

\paragraph{Models.} For QA and summarization, we generate responses using two open-source models: \textbf{LLaMA 3.3 70B} instruct~\citep{touvron2023llama} and \textbf{Granite 3.1 8B} instruct~\citep{mishra2024granite} models.
For text-to-SQL, we use few-shot prompting on a
\textbf{Codellama 34B} instruct model~\citep{rozière2024code}, a code-specialized version of Llama 2, and  \textbf{Granite 34B Code}  instruct~\citep{mishra2024granite}.

For generation, we use the following default parameters: maximum number of new tokens = 200, and for sampling-based methods, we set top-k = 20 and top-p = 0.7 when applicable. Input sequences are truncated to a maximum length of 700 tokens with padding. For consistency-based methods, we generate samples across multiple temperatures ranging from 0.25 to 1.5 in increments of 0.25, and compute average log probabilities across generated tokens for use in classification-based UQ approaches.

\paragraph{Evaluation Metrics.} 

We consider the following 3 evaluation metrics, each capturing a different facet of how confidence estimates may be utilized:  
\begin{itemize}[noitemsep,nolistsep,leftmargin=*]
\item As a performance metric, we propose \emph{accuracy from top selection} (\textbf{ATS}), which measures accuracy (fraction of correct instances) in a test set when confidences are used to select one generation from a set of generations for each query. This represents the situation where the system must provide a single response for every query and confidences are used as scores for selection among generations.
\item As a calibration metric, we choose \emph{adaptive calibrated error} (\textbf{ACE}), which bins confidence estimates into probability ranges such that each bin contains the same number of data points~\citep{nixon2019measuring}.
Formally, $ACE = \frac{1}{KB} 
\sum_{k=1}^K{\sum_{b=1}^B {|acc(b,k)-c(b,k)|}}$, where $acc(b, k)$ and $c(b, k)$ are the accuracy and confidence of adaptive calibration bin $b$ for
class label $k$. We prefer using adaptive bin sizes instead of fixed bin sizes, as the latter often results in unbalanced datapoints across bins.
We set the $\#$ of bins $B=5$ for all experiments. 
\item As a prediction metric, we consider the \emph{area under the receiver operating characteristic} (\textbf{AUROC}), which computes the area under the curve of the false positive rate vs. true positive rate when confidences are used as a probabilistic classifier for the correctness of generations. This is a standard metric that is widely used for evaluating confidence estimation. 
\end{itemize}

\begin{table*}[t!]
\caption{Comparing different UQ approaches over 3 evaluation metrics on generations from 2 models each on 3 datasets (1 per task) -- CoQA, SamSum, and Spider. Each approach is marked as either black-box (BB) or white-box (WB). Proposed approaches are listed in bold, others are baselines. Jaccard is used for all similarity-based methods. Error bars are from max. and min. values over 5 runs, each with a random $50\%$ train / $50\%$ test split. 
}
\label{table-main-3datasets}
\centering
\resizebox{0.9\linewidth}{!}{%
\begin{tabular}{l
cccccc}\toprule
Eval Metrics: 
 & ATS $\uparrow$ & ACE $\downarrow$    & AUROC $\uparrow$ &  ATS $\uparrow$ & ACE $\downarrow$  &  AUROC $\uparrow$  \\
 \toprule
 \textbf{CoQA} & \multicolumn{3}{c}{Llama3.3 70B} & \multicolumn{3}{c}{Granite3.1 8B} 
\\\cmidrule(lr){2-4}\cmidrule(lr){5-7} 
\textit{avg. log prob (WB)} & $0.27 {\scriptstyle\pm 0.01}$ & $0.272 {\scriptstyle\pm 0.019}$ & $0.72 {\scriptstyle\pm 0.02}$ & $0.06 {\scriptstyle\pm 0.01}$ & $0.808 {\scriptstyle\pm 0.002}$ & $0.86 {\scriptstyle\pm 0.03}$ \\
\textit{p(true) (BB)} & $0.19 {\scriptstyle\pm 0.02}$ & $1.195 {\scriptstyle\pm 0.012}$ & $0.36 {\scriptstyle\pm 0.01}$ & $0.03 {\scriptstyle\pm 0.00}$ & $1.004 {\scriptstyle\pm 0.006}$ & $0.60 {\scriptstyle\pm 0.03}$ \\ 
\textit{spec-ecc (BB)} & $0.12 {\scriptstyle\pm 0.02}$ & $0.331 {\scriptstyle\pm 0.018}$ & $0.21 {\scriptstyle\pm 0.02}$ & $0.06 {\scriptstyle\pm 0.01}$ & $0.685 {\scriptstyle\pm 0.010}$ & $0.19 {\scriptstyle\pm 0.02}$ \\
\textit{arith-agg (BB)} & $0.27 {\scriptstyle\pm 0.01}$ & $0.272 {\scriptstyle\pm 0.019}$ & $0.83 {\scriptstyle\pm 0.01}$ & $0.05 {\scriptstyle\pm 0.01}$ & $0.206 {\scriptstyle\pm 0.003}$ & $0.71 {\scriptstyle\pm 0.02}$ \\
\textit{clf-gen (WB)} & $0.29 {\scriptstyle\pm 0.02}$ & $0.074 {\scriptstyle\pm 0.011}$ & $0.73 {\scriptstyle\pm 0.02}$ & $0.06 {\scriptstyle\pm 0.00}$ & $0.028 {\scriptstyle\pm 0.002}$ & $0.86 {\scriptstyle\pm 0.03}$ \\
\midrule
\textit{\textbf{bayes-beta} (BB)} & $\textbf{0.38} {\scriptstyle\pm 0.02}$ & $0.093 {\scriptstyle\pm 0.010}$ & $0.90 {\scriptstyle\pm 0.01}$ & $0.17 {\scriptstyle\pm 0.01}$ & $0.028 {\scriptstyle\pm 0.001}$ & $\textbf{0.95} {\scriptstyle\pm 0.02}$ \\
\textit{\textbf{clf-pairs} (BB)} & $0.37 {\scriptstyle\pm 0.02}$ & $\textbf{0.041} {\scriptstyle\pm 0.012}$ & $\textbf{0.91} {\scriptstyle\pm 0.01}$ & $0.18 {\scriptstyle\pm 0.01}$ & $0.015 {\scriptstyle\pm 0.004}$ & $\textbf{0.95} {\scriptstyle\pm 0.02}$ \\
\textit{\textbf{clf-mean+gen} (WB)} & $0.31 {\scriptstyle\pm 0.01}$ & $0.047 {\scriptstyle\pm 0.019}$ & $0.84 {\scriptstyle\pm 0.02}$ & $0.16 {\scriptstyle\pm 0.01}$ & $0.021 {\scriptstyle\pm 0.003}$ & $0.89 {\scriptstyle\pm 0.03}$ \\
\textit{\textbf{clf-pairs+gen} (WB)} & $0.37 {\scriptstyle\pm 0.02}$ & $0.043 {\scriptstyle\pm 0.011}$ & $\textbf{0.91} {\scriptstyle\pm 0.01}$ & $\textbf{0.19} {\scriptstyle\pm 0.01}$ & $\textbf{0.014} {\scriptstyle\pm 0.002}$ & $\textbf{0.95} {\scriptstyle\pm 0.02}$ \\
\midrule
\textbf{SamSum} & \multicolumn{3}{c}{Llama3.3 70B} & \multicolumn{3}{c}{Granite3.1 8B} 
\\\cmidrule(lr){2-4}\cmidrule(lr){5-7} 
\textit{avg. log prob (WB)} & $0.51 {\scriptstyle\pm 0.03}$ & $0.419 {\scriptstyle\pm 0.025}$ & $0.57 {\scriptstyle\pm 0.02}$ & $0.15 {\scriptstyle\pm 0.02}$ & $0.656 {\scriptstyle\pm 0.016}$ & $0.40 {\scriptstyle\pm 0.01}$ \\
\textit{p(true) (BB)} & $0.50 {\scriptstyle\pm 0.03}$ & $1.099 {\scriptstyle\pm 0.015}$ & $0.50 {\scriptstyle\pm 0.00}$ & $0.46 {\scriptstyle\pm 0.02}$ & $0.911 {\scriptstyle\pm 0.013}$ & $0.50 {\scriptstyle\pm 0.01}$ \\
\textit{spec-ecc (BB)}  & $0.47 {\scriptstyle\pm 0.02}$ & $0.314 {\scriptstyle\pm 0.023}$ & $0.41 {\scriptstyle\pm 0.01}$ & $0.01 {\scriptstyle\pm 0.01}$ & $0.358 {\scriptstyle\pm 0.012}$ & $0.44 {\scriptstyle\pm 0.01}$ \\
\textit{arith-agg (BB)} & $0.53 {\scriptstyle\pm 0.02}$ & $\textbf{0.045} {\scriptstyle\pm 0.010}$ & $0.63 {\scriptstyle\pm 0.01}$ & $0.32 {\scriptstyle\pm 0.03}$ & $0.086 {\scriptstyle\pm 0.008}$ & $0.72 {\scriptstyle\pm 0.01}$ \\
\textit{clf-gen (WB)}  & $0.52 {\scriptstyle\pm 0.03}$ & $0.056 {\scriptstyle\pm 0.015}$ & $0.52 {\scriptstyle\pm 0.03}$ & $0.28 {\scriptstyle\pm 0.02}$ & $\textbf{0.050} {\scriptstyle\pm 0.007}$ & $0.62 {\scriptstyle\pm 0.02}$ \\
\midrule
\textit{\textbf{bayes-beta} (BB)} & $0.53 {\scriptstyle\pm 0.02}$ & $0.322 {\scriptstyle\pm 0.018}$ & $0.63 {\scriptstyle\pm 0.01}$ & $0.32 {\scriptstyle\pm 0.02}$ & $0.911 {\scriptstyle\pm 0.013}$ & $0.72 {\scriptstyle\pm 0.01}$ \\
\textit{\textbf{clf-pairs} (BB)} & $\textbf{0.55} {\scriptstyle\pm 0.02}$ & $0.046 {\scriptstyle\pm 0.013}$ & $\textbf{0.65} {\scriptstyle\pm 0.02}$ & $0.52 {\scriptstyle\pm 0.02}$ & $0.060 {\scriptstyle\pm 0.005}$ & $\textbf{0.86} {\scriptstyle\pm 0.01}$ \\
\textit{\textbf{clf-mean+gen} (WB)} & $0.53 {\scriptstyle\pm 0.02}$ & $0.052 {\scriptstyle\pm 0.020}$ & $0.62 {\scriptstyle\pm 0.01}$ & $0.42 {\scriptstyle\pm 0.04}$ & $0.055 {\scriptstyle\pm 0.016}$ & $0.81 {\scriptstyle\pm 0.02}$ \\
\textit{\textbf{clf-pairs+gen} (WB)} & $\textbf{0.55} {\scriptstyle\pm 0.02}$ & $\textbf{0.045} {\scriptstyle\pm 0.015}$ & $\textbf{0.65} {\scriptstyle\pm 0.02}$ & $\textbf{0.53} {\scriptstyle\pm 0.02}$ & $0.061 {\scriptstyle\pm 0.005}$ & $\textbf{0.86} {\scriptstyle\pm 0.01}$ \\
\midrule
\textbf{Spider} & \multicolumn{3}{c}{Codellama 34B} & \multicolumn{3}{c}{Granite 34B Code} 
\\\cmidrule(lr){2-4}\cmidrule(lr){5-7} 
\textit{avg. log prob (WB)} & $0.28 {\scriptstyle\pm 0.02}$ & $0.654 {\scriptstyle\pm 0.012}$ & $0.53 {\scriptstyle\pm 0.01}$ & $0.24 {\scriptstyle\pm 0.03}$ & $0.632 {\scriptstyle\pm 0.012}$ & $0.65 {\scriptstyle\pm 0.01}$ \\
\textit{p(true) (BB)} & $0.23 {\scriptstyle\pm 0.00}$ & $0.784 {\scriptstyle\pm 0.006}$ & $0.52 {\scriptstyle\pm 0.01}$ & $0.19 {\scriptstyle\pm 0.02}$ & $0.892 {\scriptstyle\pm 0.003}$ & $0.63 {\scriptstyle\pm 0.02}$ \\  
\textit{spec-ecc (BB)} & $0.01 {\scriptstyle\pm 0.00}$ & $0.201 {\scriptstyle\pm 0.006}$ & $0.37 {\scriptstyle\pm 0.01}$ & $0.02 {\scriptstyle\pm 0.00}$ & $0.551 {\scriptstyle\pm 0.009}$ & $0.20 {\scriptstyle\pm 0.01}$ \\
\textit{arith-agg (BB)} & $0.29 {\scriptstyle\pm 0.02}$ & $0.238 {\scriptstyle\pm 0.009}$ & $0.68 {\scriptstyle\pm 0.01}$ & $\textbf{0.33} {\scriptstyle\pm 0.01}$ & $0.070 {\scriptstyle\pm 0.005}$ & $0.81 {\scriptstyle\pm 0.01}$ \\
\textit{clf-gen (WB)} & $0.28 {\scriptstyle\pm 0.02}$ & $0.050 {\scriptstyle\pm 0.009}$ & $0.53 {\scriptstyle\pm 0.02}$ & $0.24 {\scriptstyle\pm 0.03}$ & $0.050 {\scriptstyle\pm 0.008}$ & $0.64 {\scriptstyle\pm 0.01}$ \\
\midrule
\textit{\textbf{bayes-beta} (BB)} & $0.29 {\scriptstyle\pm 0.02}$ & $0.298 {\scriptstyle\pm 0.018}$ & $0.68 {\scriptstyle\pm 0.01}$ & $\textbf{0.33} {\scriptstyle\pm 0.01}$ & $0.317 {\scriptstyle\pm 0.022}$ & $0.80 {\scriptstyle\pm 0.01}$ \\
\textit{\textbf{clf-pairs} (BB)} & $\textbf{0.30} {\scriptstyle\pm 0.01}$ & $0.036 {\scriptstyle\pm 0.005}$ & $\textbf{0.69} {\scriptstyle\pm 0.01}$ & $\textbf{0.33} {\scriptstyle\pm 0.02}$ & $\textbf{0.034} {\scriptstyle\pm 0.004}$ & $\textbf{0.85} {\scriptstyle\pm 0.01}$ \\
\textit{\textbf{clf-mean+gen} (WB)} & $0.29 {\scriptstyle\pm 0.02}$ & $0.036 {\scriptstyle\pm 0.005}$ & $0.67 {\scriptstyle\pm 0.01}$ & $0.31 {\scriptstyle\pm 0.02}$ & $0.039 {\scriptstyle\pm 0.002}$ & $0.80 {\scriptstyle\pm 0.01}$ \\
\textit{\textbf{clf-pairs+gen} (WB)} & $0.29 {\scriptstyle\pm 0.02}$ & $\textbf{0.034} {\scriptstyle\pm 0.006}$ & $\textbf{0.69} {\scriptstyle\pm 0.01}$ & $\textbf{0.33} {\scriptstyle\pm 0.02}$ & $0.036 {\scriptstyle\pm 0.004}$ & $\textbf{0.85} {\scriptstyle\pm 0.01}$ \\
\bottomrule
\end{tabular}
}
\end{table*}

\paragraph{Baselines.} We consider the following baselines, many of which are state-of-the-art approaches spanning all categories of UQ in LLMs:
\begin{itemize}[noitemsep,nolistsep,leftmargin=*]
\item \emph{avg. log prob} computes a probability by exponentiating the avg. logit over generated tokens;  
the generative score for sample $y_i$ is denoted $p^g_i$. 
\item \emph{spec-ecc} is a spectral clustering approach for UQ that leverages a graph Laplacian matrix computed from pairwise similarities and uses eccentricity~\citep{lin2024generating}.
\item \emph{p(true)} is
when an LLM is asked whether a generation is either True or False and the probability of the generated token (True/False) determines confidence~\cite{kadavath2022language}.
\item \emph{arith-agg} 
estimates a generation's confidence as the arithmetic mean of pairwise similarities; it is mathematically equivalent to a spectral clustering approach that uses degree~\citep{lin2024generating}. 
\item \emph{clf-gen} is a classification approach with generative score $p^g_i$ (as described above) as only feature. 
\end{itemize}
Note that we do not consider
approaches that use natural language inference for similarity~\citep{kuhn2023semantic,chen-mueller-2024-quantifying} or those requiring fine-tuning LLMs as baselines, since they are either unsuitable for structured output such as SQL or require substantial training data.

\paragraph{Proposed Methods.}

We consider the following proposed methods. Recall that $\mathbf{s}^f_i$ and $\mathbf{o}^f_i$ refer to similarity and other features respectively, for methods leveraging aggregation by classification:
\begin{itemize}[noitemsep,nolistsep,leftmargin=*]
\item \emph{bayes-beta} is the Bayesian agg. approach, where Beta distribution parameters are learned. 
\item \emph{clf-pairs} is agg. by classification when all pairwise similarities are features: $\mathbf{s}^f_i = \mathbf{s}_i$, $\mathbf{o}^f_i = \emptyset$.
\item \emph{clf-mean+gen} includes the generative score with mean similarity: $\mathbf{s}^f_i = \bar{\mathbf{s}}_i$, $\mathbf{o}^f_i = p^g_i$.
\item \emph{clf-pairs+gen} includes the generative score with all pairwise similarities:
$\mathbf{s}^f_i = \mathbf{s}_i$, $\mathbf{o}^f_i = p^g_i$.
\end{itemize}


Further experimental details about datasets, hyper-parameter choices, etc. are in Appendix~\ref{app:exp_details}.

\subsection{Main Results}

We investigate the effectiveness of our proposed classification approach using generations from $2$ different models on all datasets. 
We use a sampling procedure that generates $5$ samples each over $6$ temperatures, from $0.25$ to $1.5$ in increments of $0.25$. Evaluations are performed only on samples from the lower $3$ temperatures since the higher temperatures provide generations with lower performance. This captures the realistic scenario where the user wishes to obtain confidence estimates for only those samples they will even consider. 
We split the data randomly into half for train/test sets, and repeat the experiment $5$ times to understand variability of the results. 
To gauge the correctness of generations, we use a Rouge-L threshold of 0.5 and 0.3 for QA and summarization datasets respectively.  Jaccard is used as a similarity metric for all methods that leverage pairwise similarity.

Table~\ref{table-main-3datasets} compares various baseline and proposed UQ approaches for generations from 2 models for the CoQA, SamSum, and Spider datasets. All 3 evaluation metrics are considered -- lower ACE and higher ATS and AUROC are preferred. 
Comparing the performance
of each UQ method as shown in the rows, separately for each model, we observe that the proposed classification approaches using similarity features are generally high performing across all metrics. 
Bayesian aggregation does well on ATS and AUROC but poorly on ACE, perhaps because the conditional independence assumption that was made for tractability may not actually hold in practice.
The contrast with baselines is pronounced for CoQA where the proposed approaches are notably better.
Computing the arithmetic mean of similarities (\emph{arith-agg}) is reasonably strong for AUROC. 
Table~\ref{table-all} compares a smaller set of UQ approaches for generations on the ACE metric using generations from 
Llama-based models
for the 6 other datasets -- Natural Questions, TriviaQA, CNN Daily, XSum, BIRD, and Spider-Realistic. 
We note again that the proposed approaches perform well across datasets. 

\begin{table*}[t!]
\caption{Comparing different UQ approaches over the ACE evaluation metric on 6 datasets: Natural Questions (NQ), TriviaQA, CNN, XSum, BIRD, and Spider-Realistic. Each approach is marked as either black-box (BB) or white-box (WB). 
Proposed approaches are in bold, others are baselines. Jaccard is used for all similarity-based methods. Error bars are from max. and min. values over 5 runs, each with a random $50\%$ train / $50\%$ test split.
}
\label{table-all}
\centering
\resizebox{\linewidth}{!}{%
\begin{tabular}{lcccccc}\toprule
Dataset: & 
NQ & TriviaQA & CNN & XSum & BIRD & Spider-Realistic \\ 
\midrule
\textit{avg. log prob (WB)} & $0.913 {\scriptstyle\pm 0.000}$ & $0.914 {\scriptstyle\pm 0.002}$ &
$0.700 {\scriptstyle\pm 0.011}$ & $0.824 {\scriptstyle\pm 0.013}$ &
$0.552 {\scriptstyle\pm 0.006}$ & $0.325 {\scriptstyle\pm 0.013}$ 
\\
\textit{arith-agg (BB)} & $0.361 {\scriptstyle\pm 0.002}$ & $0.383 {\scriptstyle\pm 0.003}$ &
$0.319 {\scriptstyle\pm 0.012}$ & $0.433 {\scriptstyle\pm 0.016}$ &
$0.096 {\scriptstyle\pm 0.007}$ & $0.075 {\scriptstyle\pm 0.008}$ 
\\
\textit{clf-gen (WB)} & $\textbf{0.001} {\scriptstyle\pm 0.000}$ & $0.004 {\scriptstyle\pm 0.001}$ &
$0.052 {\scriptstyle\pm 0.013}$ & $0.038 {\scriptstyle\pm 0.017}$ &
$0.078 {\scriptstyle\pm 0.002}$ & $0.063 {\scriptstyle\pm 0.003}$ 
\\
\midrule
\textit{\textbf{clf-pairs} (BB)} & $0.005 {\scriptstyle\pm 0.010}$ & $\textbf{0.003} {\scriptstyle\pm 0.001}$ &
$\textbf{0.038} {\scriptstyle\pm 0.007}$ & $0.035 {\scriptstyle\pm 0.016}$ &
$0.089 {\scriptstyle\pm 0.007}$ & $\textbf{0.054} {\scriptstyle\pm 0.008}$ 
\\
\textit{\textbf{clf-mean+gen} (WB)} & $\textbf{0.001} {\scriptstyle\pm 0.002}$ & $\textbf{0.003} {\scriptstyle\pm 0.001}$ &
$0.049 {\scriptstyle\pm 0.017}$ & $0.035 {\scriptstyle\pm 0.019}$ &
$\textbf{0.069} {\scriptstyle\pm 0.007}$ & $0.056 {\scriptstyle\pm 0.007}$ 
\\
\textit{\textbf{clf-pairs+gen} (WB)} & $0.005 {\scriptstyle\pm 0.007}$ & $\textbf{0.003} {\scriptstyle\pm 0.001}$ &
$0.039 {\scriptstyle\pm 0.007}$ & $\textbf{0.034} {\scriptstyle\pm 0.016}$ &
$0.085 {\scriptstyle\pm 0.009}$ & $\textbf{0.054} {\scriptstyle\pm 0.005}$ 
\\
\bottomrule
\end{tabular}
}
\end{table*}

\begin{figure*}[!t]
\centering
\begin{subfigure}{0.32\textwidth}
\includegraphics[width=\textwidth]{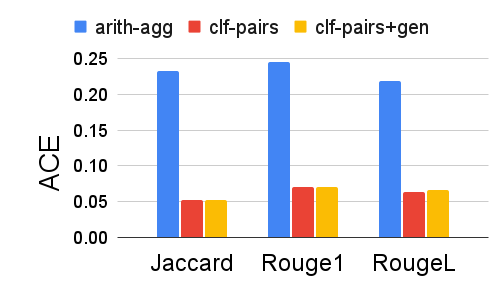}
\caption{\footnotesize CoQA}
\label{fig:ex1.1}
\end{subfigure}
\begin{subfigure}{0.32\textwidth}
\includegraphics[width=\textwidth]{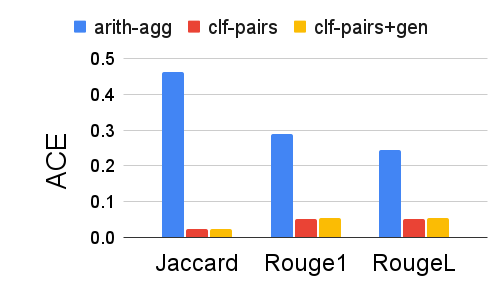}
\caption{\footnotesize SamSum}
\label{fig:ex1.2}
\end{subfigure}
\begin{subfigure}{0.32\textwidth}
\includegraphics[width=\textwidth]{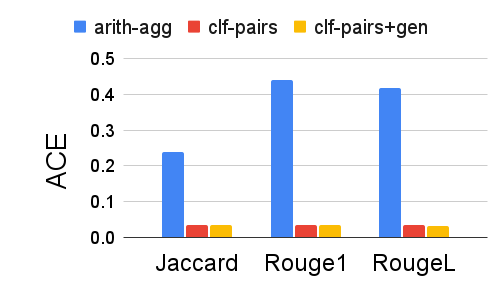}
\caption{\footnotesize Spider}
\label{fig:ex1.3}
\end{subfigure}
\caption{Effect of choice of similarity metric on the ACE metric for the CoQA, SamSum, and Spider datasets.}
\label{fig:ablation_sim_metric}
\end{figure*}

\begin{figure*}[!t]
\centering
\begin{subfigure}{0.32\textwidth}
\includegraphics[width=\textwidth]{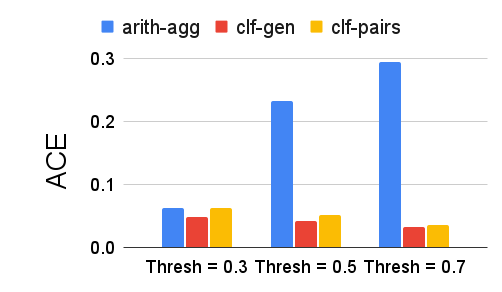}
\caption{\footnotesize Effect of Rouge-L Threshold}
\label{fig:ab1.1}
\end{subfigure}
\begin{subfigure}{0.32\textwidth}
\includegraphics[width=\textwidth]
{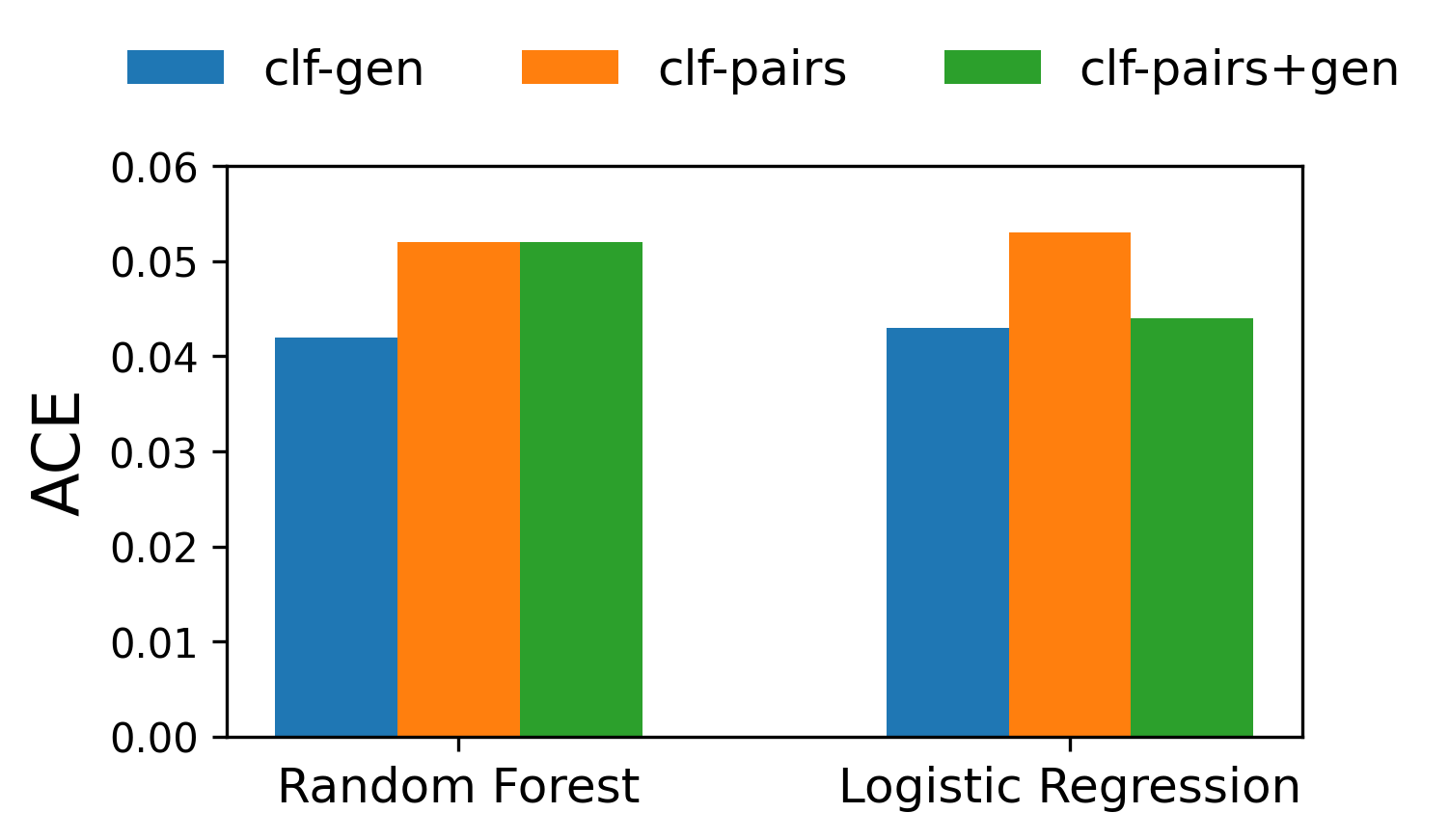}
\caption{\footnotesize Effect of Classifier}
\label{fig:ab1.2}
\end{subfigure}
\begin{subfigure}{0.32\textwidth}
\includegraphics[width=\textwidth]
{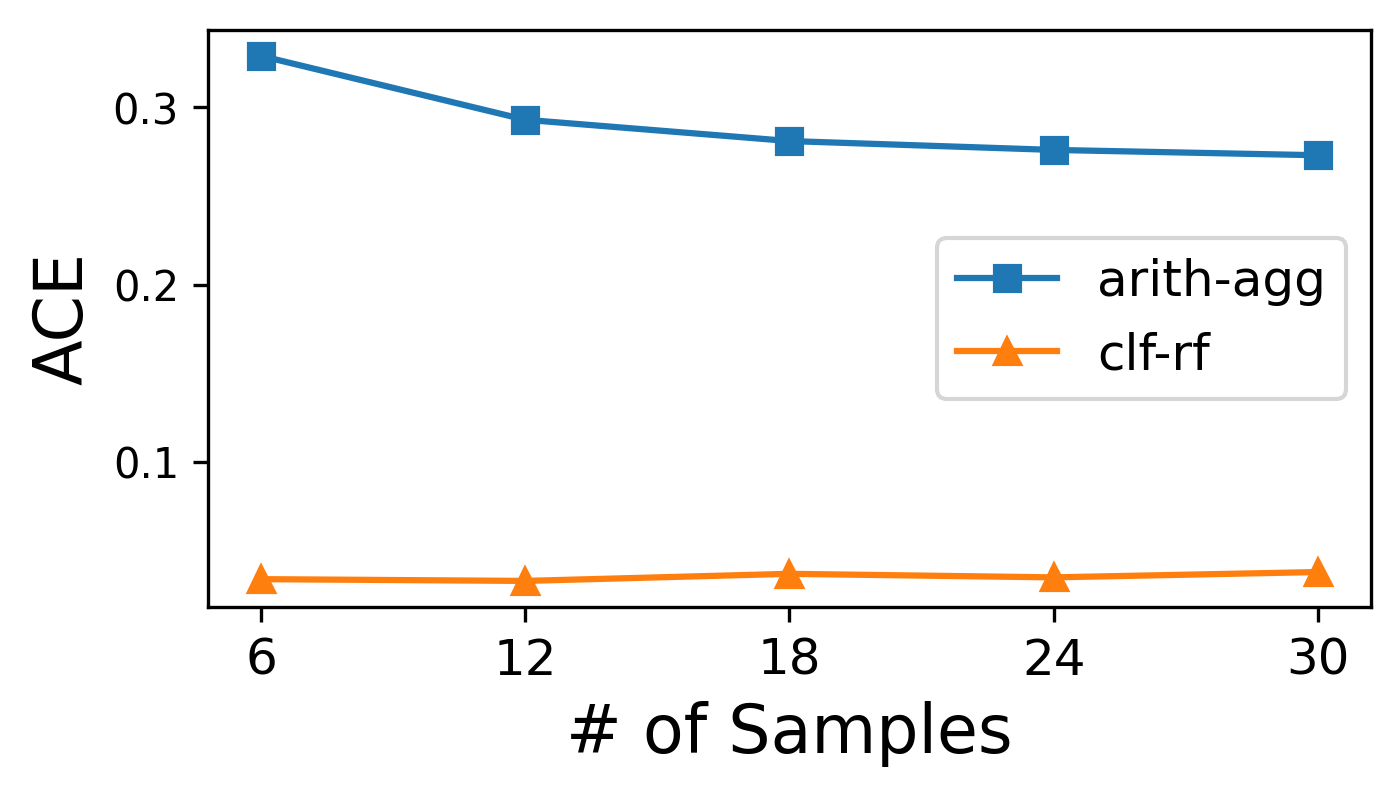}
\caption{\footnotesize Effect of \# of Samples}
\label{fig:ab1.3}
\end{subfigure}
\caption{Three different ablations on the ACE metric for the CoQA dataset. 
}
\label{fig:ablation_3diff}
\end{figure*}

\subsection{Ablations}

We conduct ablational studies to understand the impact of some our choices on the main results.



\paragraph{Similarity Metric.}

The Jaccard coefficient was used as our primary choice of similarity metric for the main experiments. Figure~\ref{fig:ablation_sim_metric} compares 3 similarity-based UQ approaches for the ACE metric on CoQA, SamSum and Spider, where we consider the Rouge-1 and Rouge-L similarity metrics in addition to Jaccard. 
The plots indicate that the trends generally remain the same regardless of choice of similarity metric 
-- the proposed classification approaches are comparable to each other and perform better than \emph{arith-agg} on ACE.





\paragraph{Three Different Ablations for CoQA.}

Each panel in Figure~\ref{fig:ablation_3diff} performs a different ablation for the ACE metric on the CoQA dataset, where we use generations from Llama 3.3 70B and similarity-based methods use Jaccard. Recall that lower ACE means better calibration.
\begin{itemize}[noitemsep,nolistsep,leftmargin=*]
\item Figure~\ref{fig:ablation_3diff}(a) explores the effect of the Rouge-L threshold, which determines whether a generation is correct. 
We compare 3 UQ approaches  (including 2 baselines) 
for the ACE metric, 
observing that trends generally remain the same across thresholds in $\{0.3, 0.5, 0.7 \}$. 
\item Figure~\ref{fig:ablation_3diff}(b) compares two types of classifiers -- random forests and logistic regression -- showing similar results.
\item Figure~\ref{fig:ablation_3diff}(c) tracks the change in ACE for two methods that use consistency as a function of the \# of samples, where the supervised method \emph{clf-pairs} is more robust than unsupervised \emph{arith-agg}. 
\end{itemize}

\paragraph{Similarity Metric and Aggregation Function for BIRD.}

In Appendix~\ref{app:sim_metric_agg_t2sql}, we conduct an extensive comparison of similarity metrics and aggregation functions for the text-to-SQL BIRD benchmark, incorporating a broader set of similarity metrics including those catering specifically to SQL generations.
Results indicate that text/token similarity metrics like Jaccard and Rouge-L are suitable for UQ even when the output is SQL, and classification with random forests is a high-performing aggregator regardless of choice of text/token metrics.




\section{Conclusion}
\label{sec:conclusion}

Assessing the confidence of LLM generations in response to a query is a crucial endeavor for enabling trusted AI. Real-world systems can benefit greatly from UQ modules that are flexible in terms of applicability to diverse generative tasks and models, and also reasonably performant for varied downstream applications that use confidence estimates.

We have proposed a general high-level similarity-based aggregation framework for UQ, leveraging pairwise similarities between multiple generated samples, 
as well as specific novel approaches within that framework.
One such approach views confidence estimation as a probabilistic classification task, where the objective is to predict the correctness of a generation using similarities with other generations for the same query 
as features.
Our methods do not rely on asking an LLM for its confidence about a generation and can be categorized as consistency-based, since they rely on using consistency between generations as a signal for confidence. 
Through an extensive empirical evaluation with $9$ datasets spanning the tasks of QA, summarization, and text-to-SQL, we show that using similarity features results in confidence estimates that fare well on various UQ evaluation metrics,  particular around minimizing calibration error.


Our data-driven methods are designed under the assumption of no data uncertainty and necessitate a small training set with samples generated in the same way for both training and testing.
Since many real-world applications often provide at least a small number of in-domain samples, and our focus in this work is specifically on no-data or low-data scenarios, we have limited our experiments to in-domain settings. Investigating out-of-domain performance with such methods is important for future work as it involves challenges such as distribution shift and domain adaptation
that warrant  separate study.

Finally, we note that results show only moderate gains over baselines on the ATS and AUROC metrics, indicating room for improvement when using the estimated confidences to distinguish between correct and incorrect responses. Further studies are needed to understand fundamental limitations of consistency-based UQ for LLMs.


\section*{Limitations}

Confidence estimation is concerned with providing confidences for LLM generations.
The consistency-based approaches proposed in this work rely on the assumption that correct generations are more similar to other generations, compared to incorrect ones. We note that while there is empirical evidence to support this assumption along with 
widespread practical adoption of similar ideas, 
further work is required to provide  theoretical justification and a more complete understanding of when the assumption may or may not hold.
Thus,
there are no guarantees associated with confidence estimates of individual instances and such approaches should be deployed with suitable caution. 



\section*{Ethical Statement}

We recognize both the positive and negative societal impacts of LLMs, including the potential misuse of our work on uncertainty quantification to lend unwarranted credibility to model outputs.
While our methods are intended to improve transparency and reliability in LLM use, we acknowledge that they could be misapplied in high-stakes contexts without proper additional safeguards. The datasets we consider are publicly available and peer-reviewed; there are no human subjects involved, and to the best of our knowledge, our work carries no direct harmful consequences.
All creators and original owners of assets have been properly credited, and licenses and terms of use have been respected. We have not conducted crowd-sourcing experiments or research with human participants. More broadly, we encourage 
continued reflection on the implications of deploying LLMs.

\section*{Acknowledgments}

We thank Nhan Pham, Kavitha Srinivas, Dharmashankar Subramanian, Long Vu, and other colleagues for their valuable comments and
feedback while discussing this work.

\bibliography{uq}

\clearpage 

\appendix

\section{Experimental Details}
\label{app:exp_details}

\subsection*{Dataset Details}

We provide additional information about the datasets considered for experiments. For QA and summarization tasks, we sub-select the first $1000$ queries in the dev/validation splits for our experimental study:

\begin{itemize}[noitemsep,nolistsep,leftmargin=*]
    \item \textbf{Question Answering Task}
    \begin{itemize}[noitemsep,nolistsep,leftmargin=*]
        \item \textbf{CoQA}~\citep{reddy2019coqa} is an open-book conversational QA dataset that measures the ability of machines to understand a text passage and answer a series of interconnected questions that appear in a conversation.
        \item \textbf{TriviaQA}~\citep{joshi2017triviaqa} is a closed-book QA dataset that is more challenging than standard QA benchmarks as the answers for a question may not be directly obtained by span prediction and the context is very long.
        \item \textbf{Natural Questions}~\citep{kwiatkowski2019nq} is also a challenging closed-book QA that contains questions from real users and requires QA systems to read and comprehend an entire Wikipedia article that may or may not contain the answer to the question.
    \end{itemize}

    \item \textbf{Summarization Task}
    \begin{itemize}[noitemsep,nolistsep,leftmargin=*]
        \item \textbf{XSum}~\citep{xsum} is a popular benchmark for abstractive summarization, focusing on generating a single-sentence summary for BBC news articles, essentially capturing their essence in one concise sentence.    
        \item \textbf{SamSum}~\citep{samsum} is a collection of human-annotated, messenger-style conversations designed for abstractive summarization research, featuring diverse styles and topics to mimic real-life dialogues. 
        \item \textbf{CNN Dailymail}~\citep{cnndailymail} is a dataset for text summarization built from human generated abstractive summary bullets generated from news stories in CNN and Daily Mail websites.
    \end{itemize}

    \item \textbf{Text-to-SQL Task}
    \begin{itemize}[noitemsep,nolistsep,leftmargin=*]
        \item \textbf{Spider}~\citep{yu2018spider} is a popular text-to-SQL benchmark covering 138 domains with 200 databases, such as academic, booking systems, and geography-related databases. The dev set has 1034 queries. 
        \item \textbf{Spider-Realistic}~\citep{deng2021structure} is a more challenging version of the Spider dev set as it modifies the natural language queries in Spider in an attempt to reflect realistic scenarios where questions do not make explicit mention of column names. It includes 508 queries.
        \item \textbf{BIRD} (BIg Bench for LaRge-scale Database Grounded Text-to-SQL Evaluation)~\citep{li2024can} is a recent cross-domain benchmark of 95 databases covering over 37 professional domains. The dev set includes 1533 queries.
    \end{itemize}
\end{itemize}

\subsection*{Model Details}

\begin{itemize}[noitemsep,nolistsep,leftmargin=*]
\item \textbf{QA and Summarization Tasks}: We generate responses using a \textbf{LLaMA 3.3 70B} instruct model~\citep{touvron2023llama} and a \textbf{Granite 3.1 8B} instruct model~\citep{mishra2024granite}.
\item \textbf{Text-to-SQL Tasks}: We use a few-shot \textbf{Codellama 34B} instruct model~\citep{rozière2024code}, which is a code-specialized version of Llama 2, trained with 500B tokens of code and code-related data, and a few-shot \textbf{Granite 34B Code} instruct model~\citep{mishra2024granite}, trained on 3-4 trillion tokens sourced from 116 programming languages.
\end{itemize}

\subsection*{Prompt Templates}

For QA datasets, we use the `prompt' field from the datasets, similar to prior work~\cite{lin2024generating}. 
Table~\ref{tab:prompt-sum-gen} shows the prompt template used for summarization datasets.
Table~\ref{tab:vanilla_prompts} shows prompt templates/examples for our few-shot approach for SQL generation with the Codellama 34B model, as well as those for a baseline (\emph{p-true}). These are illustrative and representative of prompts for other datasets for text-to-SQL.


\begin{table*}[h!]
\centering
\caption{Prompt template for summarization.}
\label{tab:prompt-sum-gen}
\begin{tabularx}{\textwidth}{X}
\toprule
\textbf{Summary Generation (Zero-Shot)} \\[0.5em]

\texttt{[INST]} \\

1. You are given an article or document. Your task is to summarize the input article in one sentence.\\
2. When generating your response, prioritize correctness, i.e., ensure that your \\
response is correct given the context and user query, and that it is grounded in the context. \\
3. Furthermore, make sure that the response is supported by the given document or context. \\

\texttt{[/INST]} \\

Summarize the following document in one sentence:\{\} \\

\bottomrule
\end{tabularx}
\end{table*}

\begin{table*}[h!]
\centering
\caption{Prompt templates for few-shot SQL generation with Codellama, as well as those for the p-true baseline.}
\label{tab:vanilla_prompts}
\begin{tabularx}{\textwidth}{X}
\toprule
\textbf{SQL Generation (Few-Shot)} \\[0.5em]

\texttt{[INST]} \\

Your task is to generate a SQL query for the given question. \\

\texttt{<<SYS>>} \\
You are given the following database schema: \{\} The SQL query must include one or more of the tables and columns from this schema. If there is only one table, do not use an alias. For multiple tables, assign aliases such as t1, t2, and prefix each column reference with the table alias (e.g., t1.age, t2.phone). The SQL query should not contain more than one table unless required by the question. Aim for efficient queries. \\[1em]
\texttt{<</SYS>>} \\[0.5em]

\textbf{Few-Shot Examples:} \\[0.5em]

Question: Show the ids and names of all documents. \\
SQL query: \texttt{SELECT document\_id, document\_name FROM Documents} \\
Question: Show the number of documents. \\
SQL query: \texttt{SELECT count(*) FROM Documents} \\
Question: Find the name and access counts of all documents, in alphabetical order of document name. \\
SQL query: \texttt{SELECT document\_name, access\_count FROM documents ORDER BY document\_name} \\
Question: Show all document ids and number of paragraphs in each document. Order by document id. \\
SQL query:  \\

\texttt{[/INST]} \\

\midrule

\textbf{p(True) (Zero-Shot)} \\[0.5em]

\textbf{Instructions:} \\

1. You are given an input question and a generated SQL query. Determine if the SQL query is correct with respect to the question. \\[0.3em]
2. Your output must be only \texttt{True} or \texttt{False}, with no extra formatting. \\[0.3em]
3. You are given the following database schema: \{\} \\[0.5em]

\textbf{True or False?} \\[0.3em]
Input: \{\} \\[0.3em]
SQL query: \{\} \\[0.3em]
Output: \\

\bottomrule
\end{tabularx}
\end{table*}

\subsection*{Details about Select Methods}

We provide some additional details about select baselines and proposed approaches below:
\begin{itemize}[noitemsep,nolistsep,leftmargin=*]
\item \emph{spec-ecc}~\citep{lin2024generating}: We apply a threshold of $0.9$ to keep only the selected eigen vectors for the spectral clustering with eccentricity baseline.
\item \emph{p-true}~\citep{kadavath2022language}: We prompt the LLM used for generations to provide their belief about whether a generation is True or False. An illustration of the zero-shot prompt template is shown in Table~\ref{tab:vanilla_prompts}. 
\item \emph{clf-?}: All classification approaches (including the baseline \emph{clf-gen}) use a random forest with a maximum depth of 4 in our experiments.
\end{itemize}

\subsection*{Computational Details}

Our UQ experiments can be conducted on a standalone CPU machine, but we use GPU machines (typically NVIDIA A100s with more than 40GB memory) for generating samples from the various LLMs as these are large models. We also access models through APIs hosted as a service but this is optional and the experiments  can be conducted on a single machine, either with GPUs or CPUs.

\begin{table*}[t]
\caption{Comparing similarity metrics and aggregation approaches using generations from a few-shot Codellama 34B model on the BIRD dev set. We include 6 similarity metrics (across both SQL and token/text categories), 2 evaluation metrics (ACE and AUROC), and 5 UQ techniques -- 2 baselines and 3 proposed methods of Bayesian aggregation with conditional Beta distributions, and aggregation by classification using logistic regression and random forests. Error bars are from max. and min. values over 5 runs, each with a random $50\%$train / $50\%$test split.}
  \label{table-sim-comp-bird}
  \centering
\resizebox{\linewidth}{!}{%
\begin{tabular}{lcccccccccl}\toprule
Eval. Metric & \multicolumn{5}{c}{ACE $\downarrow$} & \multicolumn{5}{c}{AUROC $\uparrow$} 
\\\cmidrule(lr){2-6}\cmidrule(lr){7-11}
 & \emph{spec-ecc} & \emph{arith-agg} & \emph{bayes-beta} & \emph{clf-lr} & \emph{clf-rf} & \emph{spec-ecc} & \emph{arith-agg} & \emph{bayes-beta} & \emph{clf-lr} & \emph{clf-rf} \\ \midrule
\addlinespace
\shortstack{Makiyama \\ \vspace{0.5pt}} & \shortstack{$0.652$ \\ \vspace{0.5pt} ${\scriptstyle \pm 0.010}$} & \shortstack{$0.112$ \\ \vspace{0.5pt} ${\scriptstyle \pm 0.006}$} & \shortstack{$0.171$ \\ \vspace{0.5pt} ${\scriptstyle \pm 0.066}$} & \shortstack{$0.105$ \\ \vspace{0.5pt} ${\scriptstyle \pm 0.007}$} & \shortstack{$0.109$ \\ \vspace{0.5pt} ${\scriptstyle \pm 0.007}$} & \shortstack{$0.31$ \\ \vspace{0.5pt} ${\scriptstyle \pm 0.02}$} & \shortstack{$0.69$ \\ \vspace{0.5pt} ${\scriptstyle \pm 0.02}$} & \shortstack{$0.70$ \\ \vspace{0.5pt} ${\scriptstyle \pm 0.02}$} & \shortstack{$0.70$ \\ \vspace{0.5pt} ${\scriptstyle \pm 0.02}$} & \shortstack{$0.70$ \\ \vspace{0.5pt} ${\scriptstyle \pm 0.02}$} \\
\shortstack{Output type \\ \vspace{0.5pt}} & \shortstack{$0.257$ \\ \vspace{0.5pt} ${\scriptstyle \pm 0.003}$} & \shortstack{$0.484$ \\ \vspace{0.5pt} ${\scriptstyle \pm 0.005}$} & \shortstack{$0.188$ \\ \vspace{0.5pt} ${\scriptstyle \pm 0.009}$} & \shortstack{$0.136$ \\ \vspace{0.5pt} ${\scriptstyle \pm 0.005}$} & \shortstack{$0.136$ \\ \vspace{0.5pt} ${\scriptstyle \pm 0.005}$} & \shortstack{$0.43$ \\ \vspace{0.5pt} ${\scriptstyle \pm 0.00}$} & \shortstack{$0.58$ \\ \vspace{0.5pt} ${\scriptstyle \pm 0.00}$} & \shortstack{$0.58$ \\ \vspace{0.5pt} ${\scriptstyle \pm 0.00}$} & \shortstack{$0.57$ \\ \vspace{0.5pt} ${\scriptstyle \pm 0.02}$} & \shortstack{$0.57$ \\ \vspace{0.5pt} ${\scriptstyle \pm 0.02}$} \\ \midrule
\addlinespace
\shortstack{Jaccard \\ \vspace{0.5pt}} & \shortstack{$0.648$ \\ \vspace{0.5pt} ${\scriptstyle \pm 0.010}$} & \shortstack{$0.226$ \\ \vspace{0.5pt} ${\scriptstyle \pm 0.005}$} & \shortstack{$0.126$ \\ \vspace{0.5pt} ${\scriptstyle \pm 0.006}$} & \shortstack{$0.114$ \\ \vspace{0.5pt} ${\scriptstyle \pm 0.008}$} & \shortstack{$0.093$ \\ \vspace{0.5pt} ${\scriptstyle \pm 0.008}$} & \shortstack{$0.27$ \\ \vspace{0.5pt} ${\scriptstyle \pm 0.03}$} & \shortstack{$0.76$ \\ \vspace{0.5pt} ${\scriptstyle \pm 0.01}$} & \shortstack{$0.74$ \\ \vspace{0.5pt} ${\scriptstyle \pm 0.01}$} & \shortstack{$0.75$ \\ \vspace{0.5pt} ${\scriptstyle \pm 0.02}$} & \shortstack{$\mathbf{0.78}$ \\ \vspace{0.5pt} ${\scriptstyle \pm 0.02}$} \\
\shortstack{Rouge-1 \\ \vspace{0.5pt}} & \shortstack{$0.460$ \\ \vspace{0.5pt} ${\scriptstyle \pm 0.011}$} & \shortstack{$0.388$ \\ \vspace{0.5pt} ${\scriptstyle \pm 0.005}$} & \shortstack{$0.137$ \\ \vspace{0.5pt} ${\scriptstyle \pm 0.008}$} & \shortstack{$0.097$ \\ \vspace{0.5pt} ${\scriptstyle \pm 0.007}$} & \shortstack{$0.090$ \\ \vspace{0.5pt} ${\scriptstyle \pm 0.006}$} & \shortstack{$0.31$ \\ \vspace{0.5pt} ${\scriptstyle \pm 0.01}$} & \shortstack{$0.73$ \\ \vspace{0.5pt} ${\scriptstyle \pm 0.01}$} & \shortstack{$0.77$ \\ \vspace{0.5pt} ${\scriptstyle \pm 0.01}$} & \shortstack{$0.77$ \\ \vspace{0.5pt} ${\scriptstyle \pm 0.01}$} & \shortstack{$0.77$ \\ \vspace{0.5pt} ${\scriptstyle \pm 0.01}$} \\
\shortstack{Rouge-L \\ \vspace{0.5pt}} & \shortstack{$0.505$ \\ \vspace{0.5pt} ${\scriptstyle \pm 0.010}$} & \shortstack{$0.360$ \\ \vspace{0.5pt} ${\scriptstyle \pm 0.006}$} & \shortstack{$0.149$ \\ \vspace{0.5pt} ${\scriptstyle \pm 0.011}$} & \shortstack{$0.098$ \\ \vspace{0.5pt} ${\scriptstyle \pm 0.007}$} & \shortstack{$\mathbf{0.089}$ \\ \vspace{0.5pt} ${\scriptstyle \pm 0.008}$} & \shortstack{$0.28$ \\ \vspace{0.5pt} ${\scriptstyle \pm 0.01}$} & \shortstack{$0.75$ \\ \vspace{0.5pt} ${\scriptstyle \pm 0.01}$} & \shortstack{$0.77$ \\ \vspace{0.5pt} ${\scriptstyle \pm 0.01}$} & \shortstack{$0.77$ \\ \vspace{0.5pt} ${\scriptstyle \pm 0.02}$} & \shortstack{$0.77$ \\ \vspace{0.5pt} ${\scriptstyle \pm 0.02}$} \\
\shortstack{Sbert-cos \\ \vspace{0.5pt}} & \shortstack{$0.309$ \\ \vspace{0.5pt} ${\scriptstyle \pm 0.008}$} & \shortstack{$0.557$ \\ \vspace{0.5pt} ${\scriptstyle \pm 0.006}$} & \shortstack{$0.117$ \\ \vspace{0.5pt} ${\scriptstyle \pm 0.006}$} & \shortstack{$0.099$ \\ \vspace{0.5pt} ${\scriptstyle \pm 0.009}$} & \shortstack{$0.091$ \\ \vspace{0.5pt} ${\scriptstyle \pm 0.007}$} & \shortstack{$0.40$ \\ \vspace{0.5pt} ${\scriptstyle \pm 0.01}$} & \shortstack{$0.69$ \\ \vspace{0.5pt} ${\scriptstyle \pm 0.01}$} & \shortstack{$\textbf{0.78}$ \\ \vspace{0.5pt} ${\scriptstyle \pm 0.01}$} & \shortstack{$0.77$ \\ \vspace{0.5pt} ${\scriptstyle \pm 0.01}$} & \shortstack{$0.77$ \\ \vspace{0.5pt} ${\scriptstyle \pm 0.01}$} \\
\bottomrule
\end{tabular}
}
\end{table*}



\section{Effect of Similarity Metric and Aggregation Technique on BIRD}
\label{app:sim_metric_agg_t2sql}

We conduct a more in-depth investigation into confidence estimation using similarity-based aggregation for the more complex text-to-SQL task.
Specifically, we analyze the choice of similarity metric and similarity aggregation technique using generations from a Codellama 34B model on the BIRD dataset.  
For this experiment, we generate $5$ samples each over $6$ temperatures ($\{0.25, 0.5, \cdots, 1.5 \}$), and evaluations are performed using all $6$ samples across all queries. We randomly split the data into half for train/test sets, and repeat the experiment $5$ times to study variability of the results.

We expand the set of similarity metrics to also include the following:
\begin{itemize}[noitemsep,nolistsep,leftmargin=*]
\item \emph{Embedding-based}: We include the cosine similarity between sentence BERT (sbert)~\citep{reimers-2019-sentence-bert} representations of the generations as an embedding-based metric that compares semantic similarity.
\item \emph{SQL-specific}: We also consider similarity metrics specific to SQL queries, such as the binary metric of whether two generations belong to the same SQL output type among 3 categories (simple/join/nested)~\citep{pourreza2024din}, as well as those that rely on parsing the SQL and comparing the contents of various clauses -- Aligon \citep{aligon2014similarity}, Aouiche \citep{aouiche2006clustering}, and Makiyama \citep{makiyama2015text}. Makiyama is representative and has been shown to perform well among these on a query clustering task~\citep{tang2022preqr}.
\end{itemize}

We compare a small set of aggregation functions -- 2 baselines and 3 proposed methods.
For baselines, we include two spectral clustering approaches for UQ that leverage a graph Laplacian matrix computed from pairwise similarities -- one that uses eccentricity (\emph{spec-ecc}) and another that uses degree~\citep{lin2024generating}; as mentioned previously, the latter is equivalent to simple aggregation using arithmetic mean (\emph{arith-agg}). For the proposed approaches, we include Bayesian aggregation (\emph{bayes-beta}) and classification with logistic regression (\emph{clf-lr}) as well as random forests (\emph{clf-rf}).



The rows in Table~\ref{table-sim-comp-bird} 
correspond to 6 similarity metrics and the columns correspond to $5$ UQ techniques with evaluations along 2 metrics -- ACE and AUROC.
Comparing similarity metrics, we observe that all token/text-based metrics (i.e. Jaccard, Rouge-1, Rouge-L, and Sbert-cos) generally perform well on ACE with a powerful aggregation method such as a random forest classifier. Rouge-L and sbert-cos are high performing metrics for this dataset. We also note that our proposed aggregation methods are better for calibration metrics such as ACE rather than AUROC, as sometimes they only provide marginal improvements over averaging similarities with the \emph{arith-agg} baseline.

\end{document}